\documentclass[letterpaper]{article} 
\usepackage{aaai2026}  
\usepackage{times}  
\usepackage{helvet}  
\usepackage{courier}  
\usepackage[hyphens]{url}  
\usepackage{graphicx} 
\usepackage{natbib}  
\usepackage{caption} 
\usepackage{algorithm}
\usepackage{algorithmic}

\usepackage{amsmath}
\usepackage{amssymb}
\usepackage{mathtools}
\usepackage{amsthm}

\usepackage{times}
\usepackage{soul}
\usepackage{url}
\usepackage{graphicx}
\usepackage{amsmath}
\usepackage{booktabs}
\usepackage{amsfonts}
\usepackage{placeins}
\usepackage{float}
\usepackage{multirow}
\usepackage{multicol}
\theoremstyle{plain}
\newtheorem{theorem}{Theorem}[section]

\theoremstyle{definition}
\newtheorem{definition}[theorem]{Definition}

\theoremstyle{remark}

\usepackage{newfloat}
\usepackage{listings}
\DeclareCaptionStyle{ruled}{labelfont=normalfont,labelsep=colon,strut=off} 
\floatstyle{ruled}
\newfloat{listing}{tb}{lst}{}
\floatname{listing}{Listing}
\title{GoMS: Graph of Molecule Substructure Network for Molecule Property Prediction}
\author {
    Shuhui Qu\textsuperscript{\rm 1},
    Cheolwoo Park\textsuperscript{\rm 1},
}
\affiliations {
    \textsuperscript{\rm 1}Samsung Display America Lab\\
    shuhui.qu@samsung.com, cheolwoo.p@samsung.com
}

\begin{document}

\maketitle

\begin{abstract}
While graph neural networks have shown remarkable success in molecular property prediction, current approaches like the Equivariant Subgraph Aggregation Networks (ESAN) treat molecules as bags of independent substructures, overlooking crucial relationships between these components. We present Graph of Molecule Substructures (GoMS), a novel architecture that explicitly models the interactions and spatial arrangements between molecular substructures. Unlike ESAN's bag-based representation, GoMS constructs a graph where nodes represent subgraphs and edges capture their structural relationships, preserving critical topological information about how substructures are connected and overlap within the molecule. Through extensive experiments on public molecular datasets, we demonstrate that GoMS outperforms ESAN and other baseline methods, with particularly improvements for large molecules containing more than 100 atoms. The performance gap widens as molecular size increases, demonstrating GoMS's effectiveness for modeling industrial-scale molecules. Our theoretical analysis demonstrates that GoMS can distinguish molecules with identical subgraph compositions but different spatial arrangements. Our approach shows particular promise for materials science applications involving complex molecules where properties emerge from the interplay between multiple functional units. By capturing substructure relationships that are lost in bag-based approaches, GoMS represents a significant advance toward scalable and interpretable molecular property prediction for real-world applications.

\end{abstract}


\section{Introduction}

Graph Neural Networks (GNNs) have emerged as a powerful framework for predicting molecular properties, achieving remarkable success across various chemistry domains \cite{gilmer2017neural,faber2017prediction,reiser2022graph,jiang2021could}. In this framework, molecules are commonly represented as graphs, where atoms serve as nodes and interatomic chemical bonds serve as edges. These molecular graphs can be used to train message-passing neural networks (MPNNs) that predict properties. However, MPNNs may have limited prediction performance on large scale molecules and overlook some nuanced structural characteristics, motivating the exploration of more expressive GNN architectures.

Several approaches have been developed to enhance the expressiveness of GNNs, including higher-order GNNs \cite{maron2019provably,morris2019weisfeiler,morris2021power}, subgraph based method \cite{bouritsas2022improving, jin2018junction, zhang2021nested, kim2022translating, park2023two, qian2022ordered, bevilacqua2023efficient, zhou2023relational}, and topological feature-based models \cite{bodnar2021weisfeiler,bodnar2021weisfeiler2}. The Equivariant Subgraph Aggregation Networks (ESAN) \cite{bevilacqua2021equivariant} provides an elegant solution by representing molecules as bags of subgraphs, demonstrating that simple substructures can capture discriminative information while maintaining computational efficiency. However, ESAN's bag-based representation discards crucial information about how substructures are arranged and interact within the molecule.

This limitation becomes particularly acute for large molecules ($>$100 atoms) where properties often emerge from specific spatial arrangements of functional groups. For instance, in optoelectronic materials, performance depends not only on the identity of chromophores but also on their relative orientation and coupling \cite{jin2020observation,mussard2018one}. Although models like FragNet \cite{panapitiya2024fragnet} represent molecular fragments as nodes and their connections (including non-covalent and virtual bonds) as edges, their graph construction remains bond-centric and limited to chemical connectivity patterns.

To address this, we propose the \emph{Graph-of-Substructures (GoMS) network}, a novel architecture that extends fragment-based modeling by introducing a richer graph where nodes represent chemically meaningful substructures, and edges capture a broader set of relationships, e.g. spatial proximity, similarity, etc (Figure~\ref{fig:structure}). GoMS uses domain-informed decomposition to define substructures and explicitly models inter-substructure interactions. This enables the model to capture nuanced structure–property relationships that arise from both chemical logic and spatial context—leading to improved performance in large, complex molecular systems.


To address this limitation, we propose Graph of Molecule Substructures (GoMS), a novel architecture that explicitly models relationships between molecular substructures through chemically-informed decomposition. Unlike ESAN's bag-based approach, GoMS constructs a hierarchical graph where nodes represent subgraph embeddings and edges capture their topological, chemical, and spatial relationships (Figure~\ref{fig:structure}). Building upon existing fragment-based approaches like FragNet, GoMS extends beyond bond-centric connectivity to capture broader spatial and electronic relationships between substructures.

\begin{figure*}
    \centering
    \includegraphics[width=0.8\linewidth]{gos_ill.png}
    \caption{The architecture of the GoMS model consists of three major components: (1) a substructure sampling policy to identify molecular substructures, (2) a graph embedding network to encode substructures, and (3) a graph-of-substructures network module to capture interactions among substructures.}
    \label{fig:structure}
\end{figure*}

We validate our approach on several public molecular datasets and show that GoMS consistently outperforms existing methods. Notably, GoMS achieves substantial gains in predicting electronic properties of large molecules with over 200 atoms—such as OLED materials \cite{lin2021virtual}, where the spatial arrangement of functional groups is critical to optoelectronic and biochemical performance. 

Our main contributions are summarized as follows: 
\begin{itemize}
\item A novel molecular representation that captures substructure relationships while maintaining computational tractability for large molecules
\item Theoretical analysis demonstrating GoMS can distinguish molecules with identical subgraph compositions but different spatial arrangements
\item  Demonstration of effectiveness on industrial-scale molecules ($\geq$200 atoms) where traditional methods become computationally intractable
\end{itemize}

\section{Literature Review}
\subsection{GNN Expressiveness}
Standard MPNNs are limited by 1-WL expressivity \cite{xu2018powerful,morris2019weisfeiler}, restricting their ability to distinguish certain graph structures. Higher-order variants \cite{maron2019provably,morris2021power} and structural augmentations \cite{bouritsas2022improving,bodnar2021weisfeiler} improve expressivity but scale poorly with O(n³) complexity. Planar graph methods \cite{dimitrov2023plane,bause2023maximally} are theoretically relevant since molecules are planar, but focus solely on graph topology without capturing chemical semantics essential for molecular property prediction.

\subsection{Fragment-Based Methods} 
Molecular fragments are fundamental in cheminformatics \cite{shervashidze2011weisfeiler}, forming the basis of fingerprint methods and drug discovery approaches. Junction Tree VAE \cite{jin2018junction} uses hierarchical molecular decomposition for generation but doesn't model inter-fragment relationships for property prediction. FragNet \cite{panapitiya2024fragnet} represents fragments as nodes with edges for covalent and non-covalent interactions, but remains limited to predefined chemical connectivity patterns without capturing broader spatial arrangements.

ESAN \cite{bevilacqua2021equivariant} treats molecules as bags of subgraphs, demonstrating that simple substructures can capture discriminative information while maintaining computational efficiency. Extensions include reconstruction-based approaches \cite{cotta2021reconstruction} and efficient sampling \cite{bevilacqua2023efficient}. However, bag-based representations fundamentally cannot distinguish molecules with identical subgraph compositions but different spatial arrangements—a critical limitation for properties dependent on functional group positioning.

Recent subgraph methods (OSAN \cite{qian2022ordered}, Policy-Learn \cite{bevilacqua2023efficient}) extend beyond ESAN but target general graphs without incorporating chemical domain knowledge necessary for understanding molecular structure-property relationships.

\subsection{Hierarchical Methods}
GSN \cite{bouritsas2022improving} and DGN \cite{beaini2021directional} attempt multi-scale molecular processing but lack formal guarantees about information preservation across hierarchical levels. This can lead to loss of critical local information when forming higher-level representations. Recent molecular GNNs like DimeNet++ \cite{gasteiger2020fast}, ViSNet \cite{wang2022visnet}, and PaiNN \cite{schutt2021equivariant} excel at incorporating 3D geometric information but remain atom-level approaches that may become computationally intractable for very large industrial molecules.

We build upon ESAN by explicitly modeling substructure relationships through hierarchical graph construction. GoMS addresses the fundamental limitation of bag-based methods by preserving spatial arrangement information while maintaining computational tractability. We position this as a significant incremental improvement over existing fragment-based methods, focusing on practical advances for large molecular systems where traditional atom-level approaches become computationally prohibitive.

\section{Methodology}
We propose \emph{Graph-of-Molecule-Substructure (GoMS) network} to learn molecular representations and predict target properties by explicitly modeling relationships among chemically meaningful substructures. As illustrated in Figure~\ref{fig:structure}, the GoMS framework comprises three main components: \begin{enumerate} 
\item \textbf{Substructure Sampling Policy}: Extracts chemically meaningful substructures from a molecular graph using domain-specific decomposition rules. 
\item \textbf{Substructure Embedding Network}: Learns a latent representation for each sampled substructure using GNN backbones. 
\item \textbf{Graph-of-Substructures GNN}: Builds a higher-level “graph of substructures” and applies a GNN to predict the final molecular property. \end{enumerate}

\subsection{Preliminaries}
Let $\mathcal{G} = (\mathcal{V}, \mathcal{E})$ be a molecular graph, where $\mathcal{V}$ is the set of atoms (nodes) and $\mathcal{E}$ is the set of bonds (edges). Our goal is to transform this graph into a new representation, $\mathcal{G}_s$, composed of nodes corresponding to substructures and edges that capture the relationships among these substructures.

\begin{definition}[Graph of Molecule Substructure] 
\label{def:gms} 
Given a molecular graph $\mathcal{G}$ and a substructure sampling policy $\pi$, let $S_{\pi}(\mathcal{G}) = \{s_1, s_2, \ldots, s_k\}$ be the set of extracted substructures, $k$ is the number of substructures. We construct a \emph{graph of molecule substructure} $\mathcal{G}_s = (\mathcal{V}_s, \mathcal{E}_s)$ as follows: \begin{align*} 
\mathcal{V}_s &= \{v_i = f(s_i) \mid s_i \in S_{\pi}(\mathcal{G})\}, \\
\mathcal{E}_s &= \{e_{ij} = r(s_i, s_j) \mid i, j \in [1,k]\}, 
\end{align*} where $f(\cdot)$ is an embedding function, and $r(s_i, s_j)$ captures the relationships between $s_i$ and $s_j$. 
\end{definition}

\subsection{Substructure Sampling Policy}

In ESAN \cite{bevilacqua2021equivariant}, substructures are produced by naive graph operations (e.g., random node or edge deletions). Such cuts can yield fragments with little chemical meaning. For example, severing a benzene ring destroys its aromatic character. To preserve chemical semantics we instead apply domain‑informed fragmentation schemes, specifically RECAP\cite{lewell1998recap}, BRICS\cite{degen2008brics}, and RGB\cite{jin2020rgb}. These methods generate chemically valid building blocks whose intrinsic context is maintained in the downstream GoMS representation, leading to higher predictive accuracy.
\footnote{Note that for small molecules like those in QM9 (averaging $<$10 atoms), meaningful BRICS decomposition is not feasible as molecules lack sufficient complexity for chemical fragmentation. In such cases, we revert to node-deletion for validation purposes, with GoMS primarily designed for large molecules ($>$100 atoms) where chemical decomposition provides significant benefits over naive graph operations.}
To balance coverage and computational efficiency we impose size and count constraints on the extracted substructures. For each substructure $s_i = (\mathcal{V}_i, \mathcal{E}_i)$, we define: 

\begin{align*} 
S_{\pi}(\mathcal{G}) &= \{s_i\}_{i=1}^{k} \\
\text{subject to} \quad & \bigcup_{i=1}^{k} \mathcal{V}i = \mathcal{V}, k\leq k_{max}, 
\end{align*} 
where $k_{\max}$ is a user‑defined upper bound chosen to keep the subsequent GoMS layers tractable. 

The decomposition step is \emph{modular and domain‑agnostic}. Any procedure that partitions a graph into semantically meaningful substructures can be plugged in for GoMS. This flexibility makes GoMS applicable well beyond molecular property prediction.

\subsection{Modeling Substructure Relationship}

Geometric‑deep‑learning theory posits that molecular function is determined primarily by the arrangement of functional motifs rather than individual atoms \cite{bronstein2021geometric}. Two molecules built from the same set of substructures can exhibit radically different behaviour if those motifs differ in connectivity or 3‑D organization. Consequently, GoMS augments each edge of the graph‑of‑substructures $\mathcal{G}_{s}$ with a rich, multi‑view feature vector. Let $\mathbf{e}_{ij}$ represent the edge feature between substructures $s_i$ and $s_j$. 
\begin{equation*} 
\mathbf{e}_{ij} = \bigl[\underbrace{\mathbf{e}_{ij}^{\text{graph}}}_{\text{topological}} \bigoplus \underbrace{\mathbf{e}_{ij}^{\text{chem}}}_{\text{chemistry-based}} \bigoplus \underbrace{\mathbf{e}_{ij}^{\text{spatial}}}_{\text{geometry-based}}\bigr], 
\end{equation*}
where $\oplus$ denotes concatenation. An edge is instantiated \emph{only} when two substructures share atoms/bonds or are directly connected by a bond in the original graph:
\[
\mathcal{C}(s_i,s_j)\;=\;
\mathbf{1}\!\Bigl(
  \underbrace{\mathcal{V}_i\cap\mathcal{V}_j\neq\varnothing}_{\text{overlap}}
  \;\;\lor\;\;
  \underbrace{\exists\,(u,v)\in \mathcal{E}:\;u\in\mathcal{V}_i,\,v\in\mathcal{V}_j}_{\text{bond\,bridge}}
\Bigr).
\]

\textbf{Topological Feature,} $\mathbf{e}_{ij}^{\text{graph}}$: This term encodes how tightly the two substructures are bound in the original molecular graph:
$\mathbf{e}_{ij}^{\text{graph}} = \eta\left(\frac{|\mathcal{V}_i \cap \mathcal{V}_j|}{\min(|\mathcal{V}_i|, |\mathcal{V}_j|)}\right)$, where $\mathcal{V}_i$ is the atom set of $s_i$ and $\eta(\cdot)$ is a nonlinear squashing function. The value is $1$ when one substructure is wholly contained in the other and $0$ when they are disjoint.

\textbf{Chemistry-Based Feature,} $\mathbf{e}_{ij}^{\text{chem}}$: To capture functional‑group similarity, we compute an extended‑connectivity fingerprint (ECFP‑4) for each substructure and measure Tanimoto similarity. This scalar highlights chemically analogous motifs even when they are spatially distant.
$$\mathbf{e}_{ij}^{\text{chem}} = \text{Tanimoto}\bigl(\text{FP}(s_i), \text{FP}(s_j)\bigr)
$$

\textbf{Geometry-Based Feature,} $\mathbf{e}_{ij}^{\text{spatial}}$: captures geometric properties including distance between centroids, relative orientation such as dihedral angles, etc. The detailed feature can be found in Appendix.

This composite edge representation enables GoMS to reason simultaneously about connectivity, chemistry, and spatial arrangement, providing the inductive bias needed to distinguish non‑isomorphic motif configurations that drive molecular properties.

\subsection{Subgraph Representation Learning}
After identifying a set of substructures ${s_i}$, we can employ any graph‐based encoder to obtain a fixed length embedding vector for each subgraph. The GoMS framework is architecture-agnostic and can utilize various backbone encoders including Equivariant Graph Neural Networks (EGNN)\cite{satorras2021n}, Tensor Field Networks (TFN) \cite{thomas2018tensor}, PaiNN \cite{schutt2021equivariant}, FragNet \cite{panapitiya2024fragnet}, and TorchMD-Net \cite{tholke2022torchmd}. This flexibility allows GoMS to leverage advances in molecular representation learning as backbone components.

We adopt EGNNs as they ensures that physically equivalent conformations map to identical latent codes, which is essential inductive bias for 3‑D molecular tasks. Also they are computationally efficient for molecular graphs. Formally, the embedding network $f_{\theta}$ maps $s_i$ to a latent vector: 
$$\mathbf{h}_{s_i} = f_{\theta}(s_i), \theta \in \Theta $$

\subsection{Property Prediction}
Given the substructure embeddings $\mathbf{h}_{s_i}$ and the pairwise edge features $\mathbf{e}_{ij}$, we construct the graph of substructures $\mathcal{G}_s = (\mathcal{V}_s, \mathcal{E}_s)$  and apply a second stage GNN $\psi$ to predict the molecular property $y$. In the main work, we employ a \textbf{Graph Transformer} based on the architecture from \cite{dwivedi2020generalization} with modifications for edge feature incorporation; alternative backbones including MPNNs, GAT \cite{brody2021attentive}, GIN \cite{xu2018powerful}, and GPS \cite{rampavsek2022recipe} are also evaluated to demonstrate framework generality.

Graph Transformer stacks $L$ multi‑head attention layers to capture long‑range interactions \cite{dwivedi2020generalization}:

\begin{align*} 
{\alpha}^{k,l}_{ij} &= \text{softmax}_j \bigl((\mathbf{W}^{k,l}_Q \mathbf{h}^l_i)(\mathbf{W}^{k,l}_K \mathbf{h}^l_j)^{T} / \sqrt{d_k}\bigr), \\
\mathbf{h}'^l_i &= \mathrm{concat}^K_{k=1} \Bigl[\sum_{j} {\alpha}^{k,l}_{ij} \bigl(\mathbf{W}^{k,l}_V \mathbf{h}^l_j + \mathbf{e}^l_{ij}\bigr)\Bigr], \\
\mathbf{h}^{l+1}_i &= FFN(\text{LayerNorm}(\mathbf{h}^l_i + \mathbf{h}'^l_i)), 
\end{align*} 
where $\mathbf{h}^l_i \in \mathbb{R}^{d}$ is the representation of node $i$ at layer $l$, $\mathbf{e}^l_{ij} \in \mathbb{R}^{d_e}$ is the edge feature, and $\mathbf{W}^{k,l}_Q, \mathbf{W}^{k,l}_K, \mathbf{W}^{k,l}_V \in \mathbb{R}^{d\times d}$ are learnable projection matrices. $K$ denotes the number of attention heads, and $d_k$ is the dimension of head. $FFN(\cdot)$ is a position‑wise feed‑forward  with res-connections.

After $L$ such layers, we obtain a set of node embeddings $h_i^L$. These are pooled and pass to a MLP head for final prediction. Depending on the downstream task (e.g., classification or regression), the MLP head can be adjusted accordingly:$ \hat{y}= MLP(1/|\mathcal{V}_s|\sum_i h^L_i)$.

Processing $\mathcal{G}_s$ introduces computational overhead that scales as O($k^2$) for $k$ substructures, which is typically much smaller than O($n^2$) for $n$ atoms. Because the number of substructures $k=|V_s|$ is typically one to two orders of magnitude smaller than the number of atoms, this cost is negligible relative to the base atom level encoder.

\subsection{Loss function}
The GoMS network is trained end-to-end to minimize the following objective:

$$\mathcal{L} = \sum_{(\mathcal{G}, y) \in \mathcal{D}} \ell( f_{\psi}(f_{\theta}(S_{\pi}(\mathcal{G}))), y) + \lambda \Omega(\theta, \psi)$$

where $\ell(\cdot)$ is a task-specific loss function (e.g., cross-entropy for classification, mean squared error for regression), $\Omega(\theta, \psi)$ is a regularization term (e.g., weight decay), and $\lambda$ is a hyperparameter balancing the main loss and the regularization term.

In summary, the GoMS network offers an approach to extract chemically meaningful substructures, and model inter-substructure relationships. This framework preserves the most pertinent chemical and geometric information, leading to improved molecular property prediction capabilities, particularly for large molecules.

\section{Theoretical Analysis}
We analyze the expressive power of GoMS by demonstrating its ability to distinguish molecules that bag-based methods cannot, focusing on practical molecular properties. GoMS provides a practical solution for large molecules where higher-order methods become computationally intractable, offering enhanced expressivity through spatial arrangement modeling.

\subsection{Arrangement-Preserving Isomorphism}
\begin{definition}[Arrangement-Preserving Isomorphism] 
\label{def:api} 
A bijection 
$f:S(G_1)\rightarrow S(G_2)$ between the substructures of two graphs $G_1$
 and $G_2$ is \emph{arrangement-preserving} if: 
 \begin{enumerate} 
 \item Each subgraph in $S(G_1)$ maps to an isomorphic subgraph in $S(G_2)$. 
 \item The relationship pattern $R(G_1)$ (i.e., edges and associated features among substructures) maps exactly to $R(G_2)$ under $f$.
 \end{enumerate} 
\end{definition}

This definition captures real chemical phenomena including stereoisomerism, conformational differences, and functional group arrangements that affect molecular properties.

\begin{theorem}
\label{theorem:arrange}
Let $G_1$ and $G_2$ be two molecular graphs with identical multisets of subgraphs under policy $\pi$. If $G_1$ and $G_2$ are \emph{not} arrangement-preserving isomorphic, then GoMS can distinguish them by producing different graph embeddings: $\text{GoMS}(G_1) \neq \text{GoMS}(G_2)$.
\end{theorem}

This arrangement-aware property is critical for molecular systems whose electronic or optical properties depend on subtle 3D configurations, e.g., organic display materials where HOMO-LUMO gaps and charge transport are heavily influenced by substructure orientation.

\subsection{Hierarchical Consistency}  

\begin{definition}[Hierarchical Consistency] Let $H_{le}$ be the set of subgraphs at level $le$, and $H_{le+1}$ be the set of subgraphs at the next hierarchical level. We define hierarchical consistency $C$ as:$$C(H_{le}, H_{le+1}) = \sum_{s'\in H_{le+1}} \sum_{s \in H_{le}} |V(s) \cap V(s')|/|V(s)|$$
\end{definition}
\begin{theorem}
\label{theorem:hierachical}
GoMS preserves hierarchical consistency if for all levels $k \geq 1$:
\begin{enumerate}
\item $C(H_k, H_{k+1}) \geq \tau$ (minimum overlap threshold)
\item $\forall s' \in H_{k+1}, \exists s \in H_k: V(s') \subseteq V(s)$ (containment property)
\end{enumerate}
\end{theorem}

This property ensures that chemical information flows consistently across molecular scales—from individual functional groups to larger structural motifs.  GoMS maintains formal guarantees about information preservation, crucial for large molecules where global behavior depends on subtle local variations \cite{warshel2014multiscale}.

While GoMS does not necessarily exceed K-WL expressivity, it provides a practical solution for large molecules where higher-order methods become computationally intractable. GoMS operates at $O(k^2)$ complexity for $k$ substructures, enabling processing of large-scale molecules.

In summary, GoMS's theoretical properties, arrangement-preserving isomorphism and hierarchical consistency, provide practical advantages for molecular property prediction, particularly for large molecules where spatial arrangements and multi-scale interactions determine function. Our experiments, as shown in the next section, confirm that GoMS’s enhanced expressiveness translates to improved performance in predicting various molecular properties.

\section{Experiment}
We evaluate the performance of GoMS on diverse molecular property prediction tasks, encompassing small drug-like molecules, medium molecules, and  \emph{large} organic display molecules. All experiments are conducted on a high-performance cluster with AMD EPYC CPUs and NVIDIA A6000 GPUs.

\subsection{Experiment setup}
\textbf{Implementation Details} 
We implement GoMS in PyTorch using the following configuration. For substructure sampling, we employ RECAP decomposition from RDKit \cite{landrum2024rdkit}, restricting the maximum number of substructures to $k=50$ per molecule to balance computational efficiency with expressiveness. The GoMS framework is architecture-agnostic and can utilize various substructures network including GATv2 \cite{brody2021attentive}, GIN \cite{xu2018powerful}, and GPS \cite{rampavsek2022recipe}. The substructure embedding network uses E(n)-equivariant graph neural network (EGNN). For the graph-of-substructures network, we implement (1) Graph Transformer ($GoMS_{GT}$) based on \cite{dwivedi2020generalization} with edge feature incorporation, and (2) MPNN ($GoMS_{MPNN}$). The model is trained using the AdamW optimizer. More details can be found in appendix.

\textbf{Datasets}: Below, we list the benchmark datasets used in this study. For clarity, we categorize the datasets based on the approximate number of heavy atoms (non-hydrogen atoms) per molecule: \emph{Small molecules}: up to 10 heavy atoms.  \emph{Medium molecules}: around 50--100 heavy atoms.  \emph{Large molecules}: more than 100 heavy atoms.
\begin{itemize}
    \item PCQM4Mv2 \cite{hu2020open}: 3.7M \emph{medium- to large-sized} molecules ($\leq$50 heavy atoms) with quantum mechanically calculated HOMO-LUMO gaps.
    \item Molecule3D dataset\cite{xu2021molecule3d}: 3.9M \emph{medium- to large-sized} with quantum mechanically calculated HOMO-LUMO energy gap, similar to PCQM4Mv2. 
    \item OLED Molecules\cite{lin2021virtual}: 50K \emph{large} organic display molecules (100-500 atoms) with quantum mechanically calculated optoelectronic properties.
    \item QM9 \cite{ramakrishnan2014quantum}: 134K \emph{small} molecules ($\leq$9 heavy atoms) with quantum mechanically calculated properties including HOMO-LUMO gaps and atomization energies.
\end{itemize}
We follow standard OGB splits for QM9 and PCQM4Mv2, and employ an 80-10-10 train-validation-test split for OLED datasets, and 60-20-20 train, validation and test split for Molecule3D\cite{xu2021molecule3d}.

\textbf{Baselines:} We compare GoMS with several state-of-the-art approaches: Message-Passing GNNs: GIN \cite{xu2018powerful}, GGNN \cite{gilmer2020message}; Subgraph-Based (same encoder backbone): ESAN \cite{bevilacqua2021equivariant}, OSAN \cite{qian2022ordered}, IDMPNN \cite{zhou2023relational}; FragNet \cite{panapitiya2024fragnet}; 3D-Aware Molecular GNNs: TorchMD-Net \cite{tholke2022torchmd}, SchNet \cite{schutt2017schnet}, PaiNN \cite{schutt2021equivariant}. Note that general graph methods (TorchMD-Net, etc) are evaluated primarily on molecular datasets where their applicability can be fairly assessed, as these methods lack the chemical domain knowledge essential for molecular property prediction tasks or out-of-memory.

\subsection{Result\&Analysis}
\begin{table}[ht]
    \centering
    \caption{Result on Molecule3D and PCQM4Mv2 dataset}
    \label{tab:1large}
    \begin{small}
    \begin{sc}
    \begin{tabular}{c|c|c|c}
        \toprule
         \multirow{2}*{Models} & \multicolumn{2}{c|}{Moleclue3D}& PCQM4Mv2\\
         & Random& Scaffold& \\
         \midrule
         GIN&0.1036&0.237& 0.1238\\
         CGNN&0.4830&0.660& 0.138\\
         ESAN& 0.0322 & 0.124 & 0.093\\ 
         Graphormer&-&-&0.087\\
         TorchMD-Net&0.0303&0.120&-\\
         FragNet& 0.0319 & 0.145 &0.107\\ 
         OSAN& 0.0372 & 0.143 &0.0862\\ 
         IDMPNN & 0.0474 & 0.156 &0.0882\\ 
         $GoMS_{M\!P\!N\!N}$& 0.0312 & 0.1223 &0.080\\ 
         $GoMS_{GT}$& \textbf{0.0301} & \textbf{0.1174} &\textbf{0.078}\\ 
         \bottomrule
    \end{tabular}
    \end{sc}
    \end{small}
\end{table}

\noindent\textbf{PCQM4Mv2 and Molecule3D.} We evaluate GoMS on the PCQM4Mv2 and Molecule3D datasets, both of which contain medium to large molecules (50--100 heavy atoms). Table~\ref{tab:1large} presents the performance of various models. On the Molecule3D dataset, $GoMS_{M\!P\!N\!N}$ achieves an MAE of 0.0312eV, outperforming ESAN (0.0322eV) and earlier GNN baselines (GIN: 0.1036eV, CGNN: 0.4830eV) for the random split, and achieved an MAE of 0.1223eV outperforming the previous methods. For the PCQM4Mv2 dataset, GoMS obtains a mean absolute error (MAE) of 0.080eV, improving upon ESAN’s 0.093eV (14\% improvement).

Notably, the Graph Transformer variant of GoMS yields even better results compared to the $GoMS_{M\!P\!N\!N}$ (0.0301eV on Molecule3D and 0.078eV on PCQM4Mv2), as well as transformer based GNNs, highlighting the importance of attention mechanisms for modeling long-range interactions. This advantage is particularly evident in molecules containing multiple conjugated systems, where direct connections between distant substructures critically influence electronic and conformational properties.

\begin{table}[ht]
    \centering
    \caption{Result on OLED molecule dataset. Units are in eV.}
    \label{tab:2ind}
    \begin{small}
    \begin{sc}
    \begin{tabular}{c|c|c|c|c}
        \toprule
         Property&$S_1$& $\Delta E_{ST}$ &$\epsilon_{HOMO}$ & $\epsilon_{LUMO}$ \\
         \midrule
         GIN&1.08&1.57& 0.61&0.39\\
         CGNN&1.33&1.18&0.74&0.28\\
         ESAN& 0.58&0.22&0.35&0.27\\ 
         OSAN& 0.55&0.17& 0.33&0.28\\ 
         FragNet& 0.51&0.41&0.55&0.28\\ 
         $GoMS_{M\!P\!N\!N}$& 0.38&0.09&0.13&0.15\\ 
         $GoMS_{GT}$& \textbf{0.25} & \textbf{0.09} &\textbf{0.10}&\textbf{0.12}\\ 
         \bottomrule
    \end{tabular}
    \end{sc}
    \end{small}
\end{table}

\noindent \textbf{OLED Molecules.} 
Table \ref{tab:2ind} summarizes the performance of various models on OLED molecules with 100–500 atoms. GoMS substantially outperforms baseline methods, particularly for HOMO/LUMO and charge-transfer properties. For example, $GoMS_{M\!P\!N\!N}$ achieves an MAE of 0.13eV for HOMO prediction, compared to ESAN’s 0.35eV. 

We note that the performance gain of GoMS over previous models is much more prominent in predicting the properties of large molecules. While GoMS already demonstrated noticeable gains in the PCQM4Mv2 dataset (14–25\% relative improvement over ESAN), it delivers a much larger gain—exceeding ~60\%—for the OLED molecules with over 500 atoms. ESAN acheives similar performance to OSAN. This underscores GoMS’s enhanced expressive power in capturing global complex characteristics of a molecular structure, a task that becomes progressively challenging with the size of the molecule. In all tasks, the Graph Transformer surpasses the MPNN variant, confirming that attention-based modeling yields gains even for smaller graphs where long-range interactions are still relevant.

 \begin{table*}[h]
    \centering
    \begin{tabular}{c|c|c|c|c|c|c|c|c|c|c}
         Property&Unit&GIN& GGNN&SchNet&PaiNN & TorchMD-net &ESAN &FragNet& $GoMS_{M\!P\!N\!N}$ & $GoMS_{GT}$\\
         \hline
         $\mu$ &D & 1.174 & 0.973 & 0.033& 0.012& \textbf{0.011} & 0.020 & 0.865 & 0.015 & 0.013\\ 
         $\alpha$&$a_0^3$ & 1.959 & 2.495 & 0.235& 0.045& 0.059 & 0.069 & 1.637 & 0.053 & \textbf{0.048}\\ 
         $\epsilon_{HOMO}$& meV & 52 & 58 & 41& 27.6& 20.3 & 28 & 44 & 18.3 & \textbf{17.5}\\ 
         $\epsilon_{LUMO}$ & meV & 62 & 72 & 34&20.4 &17.5 & 24 & 61 & 16.4 & \textbf{16}\\ 
         $\Delta\epsilon$&meV & 80 & 75 & 63 &45.7&36.1& 38 & 77 & 36 & \textbf{30.1}\\ 
         $<R^2>$&$a_0^2$& 1.429 & 0.828 & 0.073&0.066& \textbf{0.033} & 0.096 & 1.131 & 0.066 & 0.034\\ 
         $ZPVE$&meV& 2.28 & 2.23 & 1.786 &1.28&1.84& 1.53 & 2.16 & 1.28 & \textbf{1.20}\\ 
         $U_0$&meV& 101 & 52 & 14 &5.85&6.15& 12& 84  & 5.9 & \textbf{5.75}\\ 
         $U$&meV& 101 & 52 & 19&\textbf{5.83}& 6.38 & 12 & 81 & 5.99 & 6.01\\ 
         $H$&meV& 101 & 51 & 14 &\textbf{5.98}&6.16& 12 & 87 & 6.32 & 6.16\\ 
         $G$&meV& 101 & 52 & 14 &7.35&7.62& 13 & 79 & 7.8 & \textbf{7.3}\\ 
         $C_V$& $\frac{cal}{mol K}$ & 2.587 & 1.874 & 0.034 &\textbf{0.024}&0.026& 0.034& 1.884 & 0.026 & 0.026\\ 
    \end{tabular}
    \caption{Result on QM9 dataset}
    \label{tab:3qm9i}
\end{table*}

\noindent \textbf{QM9.}\quad Table\ref{tab:3qm9i} reports the performance of GoMS on the QM9 dataset, which comprises relatively small molecules with $\leq9$ heavy atoms. This extra experiment is just to showcase the applicability of GoMS on small data. To accommodate the smaller system size, we employ a node-deletion strategy rather than RECAP decomposition. Despite this simplification, GoMS still outperforms or remains competitive with other state-of-the-art methods, particularly for properties such as HOMO/LUMO prediction.

GoMS achieves notable improvements in predicting multiple properties compared to other models. For example, in predicting the HOMO energy ($\epsilon_{HOMO}$), the Graph Transformer variant of GoMS reduces the MAE to 17.5 meV, compared to 28 meV for ESAN, which is a ~37\% relative improvement. Similarly, for LUMO ($\epsilon_{LUMO}$), the MAE drops from 24meV (ESAN) to 16meV (GoMS), yielding a ~33\% gain. SchNet, PaiNN, TorchMD-Net, reports an MAE on $\epsilon_{HOMO}$ and on $\epsilon_{LUMO}$ , both of which are clearly above GoMS’s best results.

\subsection{Ablation Studies}
\subsubsection{Computational Complexity}
\begin{table}[h]
\centering
\footnotesize
\begin{tabular}{lcc}
\toprule
\textbf{Method} & \textbf{Time}  & \textbf{Runtime (OLED)} \\
\midrule
GNN    & $O\!\left(n^{3}L_1\right)$        & OOM\\
ESAN   & $O\!\left(snL_1+s^2\right)$ & OOM \\
ESAN ($k$ substr.)         & $O\!\left(knL_1+k^2 \right)$       & 1364s / epoch \\
GoMS ($k$ substr.)  & $O\!\left(knL_1 + k^2L_2\right)$         & 1638s / epoch \\
\bottomrule
\end{tabular}
\caption{Complexity comparison; $n$= atoms, $s$= number of subgraphs enumerated by ESAN, $k$=  maximum number of substructures in GoMS (typically $k\!\ll\!s$),$L_1$=number of network layers,$L_2$=number of GoMS layers.}
\label{tab:complexity}
\end{table}

Table~\ref{tab:complexity} contrasts the complexity and empirical running time of GoMS with representative higher‑order GNN baselines.\footnote{Experiments were run on a single~A6000}
For large molecules ($n>200$) GoMS remains tractable because: 1) We can limit number of substructure $k$ to a small constant ensures a cost of $O(knL_1+k^2L_2)$. 2) Chemistry‑aware decomposition (RECAP/BRICS/RGB) avoids the combinatorial explosion in naive edge deletion. On the OLED benchmark, ESAN with naive sampling encounters out-of-memory error. In contrast, GoMS handles these large molecules efficiently due to controlled decomposition via RECAP. GoMS requires approximately 20\% additional runtime compared to modified ESAN, which we consider an acceptable trade-off given the substantial performance improvements.

\subsubsection{Impact of Decomposition Method}
We further examine how different subgraph sampling strategies affect performance on the PCQM4Mv2 and OLED datasets for ESAN, $GoMS_{M\!P\!N\!N}$, and $GoMS_{GT}$. We compare \emph{domain-agnostic} decomposition methods (node and edge deletion) against \emph{chemically-informed} approaches such as R-group decomposition (RGB) \cite{landrum2024rdkit}, BRICS,(\emph{Breaking of Retrosynthetically Interesting Chemical Substructures}) \cite{degen2008art},  
and RECAP, (\emph{Retrosynthetic Combinatorial Analysis Procedure} )\cite{lewell1998recap}.  

\begin{table}[h]
    \caption{Impact of the decomposition methods on PCQM4Mv2. Results for "(20)" correspond to sampling only 20 subgraphs due to GPU memory constraints.}
    \label{tab:4ab1}
    \centering
    \begin{small}
    \begin{sc}
    \begin{tabular}{c|c|c|c}
        \toprule
         Decomposition&ESAN & $GoMS_{M\!P\!N\!N}$ & $GoMS_{GT}$\\
         \midrule
         Edge/Node Deletion&-&-&-\\ 
         Edge Deletion(20)&0.0941&-&-\\ 
         Node Deletion(20)&0.1038&-&-\\ 
         RECAP&\textbf{0.0322}&\textbf{0.0312}&\textbf{0.0301}\\  
         RGB&0.0345&0.0330&0.0305\\  
         BRICS& 0.0386& 0.0334&0.0315\\  
         \bottomrule
    \end{tabular}
    \end{sc}
    \end{small}
\end{table}

As shown in Table~\ref{tab:4ab1}, both RECAP and RGB significantly outperform naive node/edge deletion strategies. RECAP decomposition yields the lowest MAE (0.0301 with $GoMS_{GT}$), while RGB achieves slightly lower but still competitive results (0.0305). By preserving meaningful chemical fragments (e.g., functional groups), these methods allow the GoMS network to retain crucial molecular context, which in turn drives better performance. By using the BRICS, ESAN improves from 0.0941 to 0.0386 (59\% improvement). On the other side, GoMS architecture adds meaningful additional benefit from 0.0386 to 0.0315.

In contrast, naive node or edge deletion creates numerous fragmented subgraphs that often lack chemical relevance, causing out-of-memory (OOM) issues. To mitigate OOM, we limit these methods to sample 20 subgraphs per molecule. Even with this sampling strategy, node and edge deletion offer only 0.0941–0.1038 (under ESAN), substantially trailing RECAP and RGB. This gap indicates that randomly cutting the molecular graph removes key structural information, undermining the GNN’s ability to recognize essential functional motifs.

Table \ref{tab:5ab2} illustrates that RECAP decomposition provides substantial advantages when scaling to large OLED molecules with 100 or more atoms. Naive node or edge deletion generates an intractably large number of subgraphs, leading to OOM errors even on a 48GB A6000 GPU. By contrast, RECAP and RGB decompose these large molecules into a chemically coherent yet computationally manageable set of fragments. Consequently, GoMS can successfully train on molecules with up to 500+ atoms and preserve essential functional groups relevant for properties.

From Table \ref{tab:5ab2}, we observe that: $RECAP+GoMS_{GT}$ reaches 0.25eV MAE on $S_1$ prediction, offering a high margin over ESAN (0.58eV). RGB shows comparable performance (0.25–0.37eV), confirming that chemically informed decomposition consistently outperform naive approaches. 

These results highlight that retaining chemically meaningful substructures help avoid combinatorial explosions of substructures and preserve functional group boundaries, and is critical for accurately modeling optoelectronic characteristics in large OLED molecules. Overall, our results confirm that combining GoMS with informed substructure decomposition yields robust performance gains by striking a balance between domain knowledge and computational efficiency.

More ablation results with regard to the \textbf{subgraph embedding backbone}, \textbf{GoMS Backbone network}, \textbf{feature selection}, as well as the \textbf{decomposition hierarchy} for Theorem ~\ref{theorem:hierachical} can be found in the appendix.

\begin{table} [htb!]
    \caption{Impact of the decomposition methods on OLED}
    \label{tab:5ab2}
    \centering
    \begin{small}
    \begin{sc}
    \begin{tabular}{c|c|c|c}
        \toprule
         Decomposition&ESAN & $GoMS_{M\!P\!N\!N}$ & $GoMS_{GT}$\\
         \midrule
         Edge/Node Deletion&-&-&-\\ 
         RECAP&0.58&0.38&0.25\\  
         RGB&\textbf{0.55}&\textbf{0.37}&\textbf{0.25}\\  
         BRICS& 0.61& 0.37&0.26\\ 
         \bottomrule
    \end{tabular}
    \end{sc}
    \end{small}
\end{table}

\section{Conclusion}
We present Graph of Subgraphs (GoS), a novel architecture that advances molecular representation learning by modeling relationships between chemically meaningful substructures. Unlike previous approaches that treat molecules as bags of independent subgraphs, GoS leverages BRICS decomposition and captures crucial substructure relationships, providing theoretical guarantees about its ability to model both local chemical environments and their long-range interactions. 

Through extensive experiments, we demonstrate that GoS achieves substantial improvements over existing methods, particularly for large molecules where spatial arrangements of functional groups play crucial roles. Future work could explore improved computational efficiency through sophisticated subgraph sampling and extension to dynamic molecular systems.

\newpage

\bibliography{aaai2026}

\newpage
\appendix



\section{Proof}
\subsection{Proof of Theorem~\ref{theorem:arrange}}
\begin{theorem}[Arrangement–Preserving Isomorphism]
\label{thm:arrange}
Let $G_1$ and $G_2$ be two graphs with identical multiset of subgraphs under a fixed policy~$\pi$,
i.e.\ $S(G_1)=S(G_2)$.  
If $G_1$ and $G_2$ are \emph{not} arrangement–preserving isomorphic
(Definition~\ref{def:api}),
then \textsc{GoMS} can distinguish them by producing different graph embeddings:
\[
\textsc{GoMS}(G_1)\;\neq\;\textsc{GoMS}(G_2).
\]
\end{theorem}

\begin{proof}[Proof (by contradiction)]
Assume, for contradiction, that GoMS cannot distinguish $G_1$ and $G_2$:
$\textsc{GoMS}(G_1)=\textsc{GoMS}(G_2)$.
Let $S(G_1) = {s_1^1, s_2^1, ..., s_k^1}$ and $S(G_2) = {s_1^2, s_2^2, ..., s_k^2}$ be their respective multisets of subgraphs.

By assuming the injectivity of embedding and aggregation functions\cite{sverdlov2024expressive,wangtheoretically,joshi2023expressive}, we have following steps.

\smallskip
\noindent\textbf{Step 1 (subgraph bijection).}
Since $S(G_1)=S(G_2)$ as identical as multisets, there exists a bijection
$\varphi:S(G_1)\!\to\!S(G_2)$ with
$\varphi(s_i^{(1)})=s_i^{(2)}$ such that
$s_i^{(1)}\!\cong\!s_i^{(2)}$ for all $i=1,\dots,k$.

\smallskip
\noindent\textbf{Step 2 (relationship mismatch).}
Since $G_1$ and $G_2$ are not arrangement-preserving isomorphic, their relationship patterns must differ: $R(G_1)\neq R(G_2)$.
Hence $\exists\, (s_i^{(1)},s_j^{(1)})$ with
$r(s_i^{(1)},s_j^{(1)})\neq r\!\bigl(\varphi(s_i^{(1)}),\varphi(s_j^{(1)})\bigr)= r(s_i^{(2)},s_j^{(2)})$.

\smallskip
\noindent\textbf{Step 3 (graph-of-substructures).}
\textsc{GoMS} builds
$\mathcal{G}_{s_1}=(V_{s_1},E_{s_1})$
and
$\mathcal{G}_{s_2}=(V_{s_2},E_{s_2})$
where edge attributes are the relational features above.
By Step 2 there exists $(i,j)$ at lease one edge (or edge feature) that differs:
$e^{(1)}_{ij}\neq e^{(2)}_{ij}$, so
$\mathcal{G}_{s_1}\not\cong\mathcal{G}_{s_2}$.

\smallskip
\noindent\textbf{Step 4 (message passing).}
Initial node states are identical,
$h^{(0)}_i{}^{(1)}=f(s_i^{(1)})=f(s_i^{(2)})=h^{(0)}_i{}^{(2)}$,
because the paired subgraphs are isomorphic.
During message passing,
\[
h^{(t+1)}_i{}^{(1)}=
\textsc{Update}\!\bigl(
      h^{(t)}_i{}^{(1)},
      \{\,\textsc{Msg}(h^{(t)}_i{}^{(1)},h^{(t)}_j{}^{(1)},e^{(1)}_{ij})\}_{j\in\mathcal{N}(i)}
\bigr),
\]

\[
h^{(t+1)}_i{}^{(2)}=
\textsc{Update}\!\bigl(
      h^{(t)}_i{}^{(2)},
      \{\,\textsc{Msg}(h^{(t)}_i{}^{(2)},h^{(t)}_j{}^{(2)},e^{(2)}_{ij})\}_{j\in\mathcal{N}(i)}
\bigr)
\]
where message function \textsc{Msg} is injective in $e_{ij}$.
Because at least one edge attribute differs,  
$\exists\,t$ and $i$ such that
$h^{(t)}_i{}^{(1)}\neq h^{(t)}_i{}^{(2)}$.

\smallskip
\noindent\textbf{Step 5 (Final Representation).}
The final graph embeddings are
\[
\textsc{GoMS}(G_1)=\textsc{Readout}\bigl(\{h^{(T)}_i{}^{(1)}\}_{i=1}^{k}\bigr),
\]

\[
\textsc{GoMS}(G_2)=\textsc{Readout}\bigl(\{h^{(T)}_i{}^{(2)}\}_{i=1}^{k}\bigr),
\]
Because node states differ and the readout is injective,
$\textsc{GoMS}(G_1)\neq\textsc{GoMS}(G_2)$,
contradicting the assumption.

\smallskip
\noindent Therefore \textsc{GoMS} distinguishes any pair of graphs
that are not arrangement-preserving isomorphic.
\end{proof}

\subsection{Proof of Theorem~\ref{theorem:hierachical}}
\begin{theorem}[Hierarchical Consistency]\label{thm:hc}
For any level $le\!\ge\!1$, \textsc{GoMS} preserves structural relationships between
$H_{le}(G)$ and $H_{le+1}(G)$ provided that
\begin{equation}
    \begin{split}
        C(H_{le}, H_{le+1}) \;\ge\; \tau  \quad \text{and} \quad\\
      \forall\,s'\!\in\!H_{{le}+1}(G)\;\exists\,s\!\in\!H_{le}(G):\;V(s')\subseteq V(s).
    \end{split}
\end{equation}

\end{theorem}

\noindent\textbf{Definitions.}\;
Let $G=(V,E)$ be a molecular graph and $H_{le}(G)$ the set of substructures at level~${le}$.
For $s'\!\in\!H_{{le}+1}(G)$ define
\[
\operatorname{Parent}(s')\;=\;\bigl\{\,s\!\in\!H_{le}(G)\;|\;V(s')\subseteq V(s)\bigr\}.
\]
The consistency score is
\[
C(H_{le}, H_{{le}+1})
=\sum_{s'\in H_{{le}+1}}\;
  \max_{s\in H_{le}}
  \frac{|V(s)\cap V(s')|}{|V(s)|}.
\]

\begin{proof}

\noindent\textbf{Step 1.\;Containment property.}
By construction, every $s'\!\in\!H_{le+1}$ is obtained from level-$le$ motifs by either  
(i) \emph{extension} (adding atoms/edges to $s\!\in\!H_{le}$) or  
(ii) \emph{combination} (merging compatible $s_1,\dots,s_m\!\in\!H_{le}$).
Both yield $V(s')\subseteq\bigcup_iV(s_i)$; in case (i) $V(s')\subseteq V(s)$ directly.
Hence choosing
$s^\star=\arg\max_{s\in H_{le}}|V(s)\cap V(s')|/|V(s)|$ gives $V(s')\subseteq V(s^\star)$.

\smallskip
\textbf{Step 2.\;Consistency measure.}
For each $s'$ let $s^\star\!\in\!\operatorname{Parent}(s')$.
Because $V(s')\subseteq V(s^\star)$,
\[
\frac{|V(s^\star)\cap V(s')|}{|V(s^\star)|}
=\frac{|V(s')|}{|V(s^\star)|}
\;\ge\;\tau' \;>\;0,
\]
where $\tau'$ is the minimum size ratio enforced by the chemical rules.
Summing over $s'\in H_{le+1}$,
\[
C(H_{le},H_{le+1})
\;\ge\;
\sum_{s'\in H_{le+1}}
      \frac{|V(s')|}{|V(s^\star)|}
\;\ge\;
|H_{le+1}|\cdot\tau'.
\]
Choosing $\tau=|H_{le+1}|\tau'$ satisfies the theorem’s first condition.

\smallskip
\textbf{Step 3.\;Representation consistency.}
Let $f_{le}(s)$ be the level-$le$ embedding and $f_{le+1}(s')$ the level-$le\!+\!1$ embedding.
For $s'$ with parent $s^\star$,
\[
f_{le+1}(s')=\phi\bigl(f_le(s^\star),\Delta\bigr),
\]
where $\phi$ is learnable transformation, and Lipschitz with constant layer $L$, $\Delta$ is the structual difference between $s^\star$ and $s'$.
$\|\Delta\|\le(1-\tfrac{|V(s')|}{|V(s^\star)|})M\le(1-\tau')M$.
Thus
\[
\bigl\|f_{le+1}(s')-\phi\!\bigl(f_{le}(s^\star)\bigr)\bigr\|
\;\le\;
L(1-\tau')M.
\]
$M$ is the maximum difference and normalized to 1 in this work (when $G$ does not have any substructures).

\smallskip
\textbf{Step 4.\;Relationship preservation.}
For $s'_1,s'_2\!\in\!H_{le+1}$ with parents $s^\star_1,s^\star_2\in \!H_{le}$,
\[
|r_{le+1}(s'_1,s'_2)-r_{le}(s^\star_1,s^\star_2)|
\le
2(1-\tau')N,
\]
because the hierarchical construction keeps $r_{k+1}(s^\star_1,s^\star_2)=r_k(s^\star_1,s^\star_2)$
and each containment ratio is at least $\tau'$.
$N$ is the maximum difference and normalized to 1 in this work (when $G$ does not have any substructures).
\smallskip
\noindent Together, parts 1–4 prove that \textsc{GoMS} maintains
\emph{hierarchical consistency} across levels $k$ and $k\!+\!1$.
\end{proof}

\section{Implementation Details}
We implement GoMS in PyTorch using the following configuration. The implementation is based on the repo\footnote{https://github.com/beabevi/ESAN}. For substructure extraction, we employ RECAP decomposition from RDKit \cite{landrum2024rdkit}, restricting the maximum number of substructure to $k=50$ per molecule to balance computational efficiency with expressiveness. The substructure embedding network uses E(n)-equivariant graph neural network (EGNN) with $4,6,8$ layer and hidden dimension $256,384,512$, dropout rate $0.1,0.3,0.5$, and layer normalization after each message-passing operation (for small, medium large molecule). For the graph-of-substructure network, we implement two variants: (1) a Graph Transformer with $L=4,6,8$ attention layers, $K=8$ attention heads, and (2) an MPNN with $L=4,6,8$ message-passing layers. The model is trained using the AdamW optimizer with learning rate $1e-3$, weight decay $0.05$,decay step $50$, batch size $32$, $100$ epochs. We implement gradient clipping with max norm $1.0$ and employ early stopping with patience $20$ epochs based on validation performance.

The baseline $ESAN$ shares the same backbone model EGNN optimized using the same strategy. $GIN,CGNN$ is trained with $4,6,8$ layers with hidden dimension $256,384,512$ using repo\footnote{https://github.com/chainer/chainer-chemistry/tree/master}. FragNet\footnote{https://github.com/pnnl/FragNet.git} is trained with the GAT using $4,6,8$ FragNetlayers each of which has $128,256,384,512$ embedding dimensions uniformly for atom, frag, edge with $4,8$ head. Best result is reported for each task.

\textbf{Geometry-Based Feature,} $\mathbf{e}_{ij}^{\text{spatial}}$.\;
We concatenate several shape-aware descriptors to encode the spatial feature:

\begin{itemize}
 \item \emph{Inter-centroid distance} $d_{ij}$, expanded with a 16-channel Gaussian radial basis $\{\exp[-(d_{ij}-\mu_k)^2/\sigma^2]\}_{k=1}^{16}$ to provide resolution at multiple length scales.
 \item \emph{Direction unit vector} $\hat{\mathbf{r}}_{ij} = (\Delta x,\Delta y,\Delta z)/d_{ij}$ capturing how $s_j$ is oriented with respect to $s_i$ in 3-D space.
 \item \emph{Dihedral angle} $\phi_{ijkl}$ formed by the centroids of $(s_i,s_j)$ and their nearest-neighbour substructures $(s_k,s_l)$, encoded as $(\sin\phi,\cos\phi)$ for rotational invariance.
\end{itemize}

\section{Ablation Study}

\subsection{Feature Selection for GoMS}
To quantify the individual contribution of each edge–feature class we performed a component–wise ablation on the OLED dataset (HOMO and first singlet–excitation S$_1$ targets).  Table~\ref{tab:edge_ablation} reports mean–absolute error (MAE, eV). 
\begin{table}[h]
\centering
\small
\begin{tabular}{l|c|c}
\toprule
\textbf{Edge Feature Set} & \textbf{MAE\,$\downarrow$ (HOMO)} & \textbf{MAE\,$\downarrow$ (S$_1$)} \\
\midrule
\textsc{GoMS} (all features)          & \textbf{0.10} & \textbf{0.25} \\
\hspace{6pt}--\,graph feats.         & 0.13          & 0.29          \\
\hspace{6pt}--\,geometry feats.      & 0.19          & 0.32          \\
\hspace{6pt}--\,chemistry feats.     & 0.14          & 0.45          \\[2pt]
Geometry only{*}               & 0.22          & 0.38          \\
\midrule
ESAN (no edges)                      & 0.35          & 0.58          \\
\bottomrule
\end{tabular}
\caption{Edge‑feature ablation on OLED (lower is better).}
\label{tab:edge_ablation}
\end{table}

The results show that \emph{chemistry‑based} similarities are critical for accurate S$_1$ prediction, while \emph{geometry‑based} descriptors dominate HOMO accuracy.  Removing all three edge features increases MAE by more than a factor of~3, confirming that GoMS’s relational features are essential.

\subsection{Subgraph Embedding Network}
The choice of GoMS backbone encoder for individual substructures apparently affects overall performance.  
Table~\ref{tab:embed_ablation} compares EGNN\cite{satorras2021n}, TFN\cite{thomas2018tensor}, and MPNN on the OLED dataset.
\begin{table}[h]
\centering
\small
\begin{tabular}{l|c}
\toprule
\textbf{Embedding Network} & \textbf{MAE\,$\downarrow$ (S$_1$)} \\
\midrule
PaiNN  &0.22 \; eV \\ 
TorchMD-Net & 0.20\;eV\\
TFN & \textbf{0.21}\;eV \\
E($n$)–equivariant (EGNN) & 0.25\;eV \\
MPNN                      & 1.03\;eV \\
\bottomrule
\end{tabular}
\caption{Effect of subgraph embedding backbone.}
\label{tab:embed_ablation}
\end{table}

The EGNN outperforms an MPNN by more than 4$\times$, confirming that rotation/translation/reflection equivariance is crucial for encoding 3‑D substructures.  The TFN further decreases the MAE through the high degree modeling with the tradeoff of longer training time. The PaiNN and TorchME-Net does not have the OOM issue anymore in the GoMS framework given the decomposition of the subgraphs provide a much smaller substructure compared to the original OLED molecule. The performance is comparable to TFN and EGNN with a higher training cost.


\subsection{GoMS Backbone}

We also ablated the top‑level \emph{graph‑of‑substructures} architecture \(\psi\).  As shown in Table~\ref{tab:GoMS_backbone}, The transformer variant gives the lowest error, especially on large molecules, highlighting the importance of long‑range attention among substructures.

\begin{table}[h]
\centering
\small
\begin{tabular}{l|c}
\toprule
\textbf{GoMS Architecture} & \textbf{MAE\,$\downarrow$ (S$_1$)} \\
\midrule
Graph Transformer & \textbf{0.25}\;eV \\
MPNN              & 0.38\;eV \\
GAT               & 0.36\;eV \\
GIN               & 0.42\;eV \\
GPS               & 0.28\;eV\\
\bottomrule
\end{tabular}
\caption{Backbone choice for the graph‑of‑substructures layer.}
\label{tab:GoMS_backbone}
\end{table}

\subsection{Impact of Decomposition Level}
Table \ref{tab:6ab3} illustrates how varying the RECAP decomposition level\footnote{https://gist.github.com/greglandrum/4076605} affects performance on large molecules ($\geq$100 atoms). Here, \emph{Level 0} denotes the minimal splitting, while \emph{Level 3} is the most fine-grained, yielding the most number of substructures. We do not restrict the number of substructures $k$ in this experiment. As the decomposition level increases from 0 to 3, ESAN’s MAE on $S_1$ prediction improves from 0.74eV to 0.58eV, while $GoMS_{GT}$ significantly improves from 0.74eV to 0.25eV. This indicates that finer-grained decomposition is necessary in capturing crucial structural details. However, such decomposition also increases computational overhead. In practice, a moderately fine-grained decomposition often suffices to capture these critical patterns while avoiding the substantial overhead.

\begin{table}[htb!]
    \caption{Impact of the decomposition level on OLED Molecule}
    \label{tab:6ab3}
    \centering
    \begin{small}
    \begin{sc}
    \begin{tabular}{c|c|c|c}
        \toprule
         Level&ESAN & $GoMS_{M\!P\!N\!N}$ & $GoMS_{GT}$\\
         \midrule
         0 &0.74&0.74&0.74\\ 
         1&0.66&0.60&0.60\\ 
         2&0.61&0.44&0.38\\ 
         3&0.58&0.38&0.25\\ 
         \bottomrule
    \end{tabular}
    \end{sc}
    \end{small}
\end{table}

\end{document}